\pdfoutput=1

\documentclass[11pt,a4paper]{article}
\usepackage[hyperref]{acl2021}
\usepackage{times}
\usepackage{latexsym}

\usepackage{microtype}

\usepackage{amsmath}
\usepackage{booktabs}
\usepackage{graphicx}

\usepackage{tikz}

\aclfinalcopy
\usepackage{amsmath}
\usepackage{tabularx}
\usepackage{booktabs}
\usepackage{amssymb}
\usepackage{pifont}
\usepackage{bm}
\usepackage{multirow}
\usepackage{tcolorbox}
\usepackage[framemethod=tikz]{mdframed}
\usepackage{tikz,pgfplots}
\usepgfplotslibrary{groupplots}
\pgfplotsset{compat=1.13}

\newcommand{\V}[1]{\mathbf{#1}}

\definecolor{g-red}{HTML}{DB4437}
\definecolor{g-blue}{HTML}{4285F4}
\definecolor{g-green}{HTML}{0F9D58}
\definecolor{g-yellow}{HTML}{F4B400}
\definecolor{g-orange}{HTML}{FF9800}
\definecolor{g-grey}{HTML}{9E9E9E}
\definecolor{uw}{RGB}{138,43,226}
\definecolor{stanford}{RGB}{255,69,0}
\definecolor{const}{RGB}{68, 110, 182}
\definecolor{head}{RGB}{246, 180, 32}
\definecolor{freq}{RGB}{0, 0, 0}

\newmdenv[innerlinewidth=0.5pt, roundcorner=4pt,linecolor=black,innerleftmargin=6pt,
innerrightmargin=6pt,innertopmargin=6pt,innerbottommargin=6pt]{examplebox}

\title{Coreference Resolution without Span Representations}
\author{Yuval Kirstain\thanks{\;\;Equal contribution.}$\ \hspace{0.05cm}$~~~~Ori Ram$^{*}$~~~~Omer Levy\\ \\
\  Blavatnik School of Computer Science, Tel Aviv University \\
\texttt{\{yuval.kirstain,ori.ram,levyomer\}@cs.tau.ac.il}
}

\begin{document}
\maketitle

\begin{abstract}
The introduction of pretrained language models has reduced many complex task-specific NLP models to simple lightweight layers.
An exception to this trend is coreference resolution, where a sophisticated task-specific model is appended to a pretrained transformer encoder.
While highly effective, the model has a very large memory footprint -- primarily due to dynamically-constructed span and span-pair representations -- which hinders the processing of complete documents and the ability to train on multiple instances in a single batch.
We introduce a lightweight end-to-end coreference model that removes the dependency on span representations, handcrafted features, and heuristics.
Our model performs competitively with the current standard model, while being simpler and more efficient.
\end{abstract}

\section{Introduction}

Until recently, the standard methodology in NLP was to design task-specific models, such as BiDAF for question answering \cite{Seo2017BidirectionalAF} and ESIM for natural language inference \cite{chen-etal-2017-enhanced}.
With the introduction of pretraining, many of these models were replaced with simple output layers, effectively fine-tuning the transformer layers below to perform the traditional model's function \cite{gpt1}.
A notable exception to this trend is \textit{coreference resolution}, where a multi-layer task-specific model \cite{lee-etal-2017-end, lee-etal-2018-higher} is appended to a pretrained model \cite{joshi-etal-2019-bert, joshi-etal-2020-spanbert}. 
This model uses intricate span and span-pair representations, a representation refinement mechanism, handcrafted features, pruning heuristics, and more. 
While the model is highly effective, it comes at a great cost in memory consumption, limiting the amount of examples that can be loaded on a large GPU to a single document, which often needs to be truncated or processed in sliding windows. 
Can this coreference model be simplified?

We present \textit{start-to-end} (s2e) coreference resolution: a simple coreference model that does \textit{not} construct span representations.
Instead, our model propagates information to the span boundaries (i.e., its start and end tokens) and  computes mention and antecedent scores through a series of bilinear functions over their contextualized representations.
Our model has a significantly lighter memory footprint, allowing us to process multiple documents in a single batch, with no truncation or sliding windows.
We do not use any handcrafted features, priors, or pruning heuristics.

Experiments show that our minimalist approach performs on par with the standard model, despite removing a significant amount of complexity, parameters, and heuristics.
Without any hyperparameter tuning, our model achieves 80.3 F1 on the English OntoNotes dataset \cite{pradhan-etal-2012-conll}, with the best comparable baseline reaching 80.2 F1 \cite{joshi-etal-2020-spanbert}, while consuming less than a third of the memory.
These results suggest that transformers can learn even difficult structured prediction tasks such as coreference resolution without investing in complex task-specific architectures.\footnote{Our code and model are publicly available: \url{https://github.com/yuvalkirstain/s2e-coref}}

\section{Background: Coreference Resolution}
\label{sec:background}

Coreference resolution is the task of clustering multiple mentions of the same entity within a given text.
It is typically modeled by identifying entity mentions (contiguous spans of text), and predicting an \textit{antecedent} mention $a$ for each span $q$ (query) that refers to a previously-mentioned entity, or a null-span $\varepsilon$ otherwise. 



\citet{lee-etal-2017-end,lee-etal-2018-higher} introduce \textit{coarse-to-fine} (c2f), an end-to-end model for coreference resolution that predicts, for each span $q$, an antecedent probability distribution over the candidate spans $c$:
\begin{align*}
P \left( a = c | q  \right) = \frac{\exp(f(c, q))}{\sum_{c^\prime} \exp(f(c^\prime, q))}
\end{align*}
Here, $f(c, q)$ is a function that scores how likely $c$ is to be an antecedent of $q$. This function is comprised of mention scores $f_m(c), f_m(q)$ (i.e. is the given span a mention?) and a separate antecedent score $f_a(c, q)$:
\begin{align*}
f(c, q) = 
\begin{cases}
f_m(c) + f_m(q) + f_a(c, q) & c \neq \varepsilon \\
0 & c = \varepsilon
\end{cases}
\end{align*}
Our model (Section~\ref{sec:model}) follows the scoring function above, but differs in how the different elements $f_m(\cdot)$ and $f_a(\cdot)$ are computed. We now describe how $f_m$ and $f_a$ are implemented in the c2f model.

\paragraph{Scoring Mentions}
In the c2f model, the mention score $f_m(q)$ is derived from a vector representation $\V{v}_q$ of the span $q$ (analogously, $f_m(c)$ is computed from $\V{v}_c$).
Let $\V{x}_i$ be the contextualized representation of the $i$-th token produced by the underlying encoder.
Every span representation is a concatenation of four elements: the representations of the span's start and end tokens $\V{x}_{q_s}, \V{x}_{q_e}$, a weighted average of the span's tokens $\hat{\V{x}}_q$ computed via self-attentive pooling, and a feature vector $\phi(q)$ that represents the span's length:
\begin{align*}
\V{v}_q = [\V{x}_{q_s}; \V{x}_{q_e}; \hat{\V{x}}_q; \phi(q)]
\end{align*}
The mention score $f_m(q)$ is then computed from the span representation $\V{v}_q$:
\begin{align*}
f_m(q) = \V{v}_m \cdot \text{ReLU}(\V{W}_m \V{v}_q)
\end{align*}
where $\V{W}_m$ and $\V{v}_m$ are learned parameters. Then, span representations are enhanced with more global information through a refinement process that interpolates each span representation with a weighted average of its candidate antecedents.
More recently, \citet{xu-choi-2020-revealing} demonstrated that this span refinement technique, as well as other modifications to it (e.g. entity equalization \cite{kantor-globerson-2019-coreference}) do not improve performance. 

\paragraph{Scoring Antecedents}
The antecedent score $f_a(c, q)$ is derived from a vector representation of the span \textit{pair} $\V{v}_{(c, q)}$.
This, in turn, is a function of the individual span representations $\V{v}_c$ and $\V{v}_q$, as well as a vector of handcrafted features $\phi(c, q)$ such as the distance between the spans $c$ and $q$, the document's genre, and whether $c$ and $q$ were said/written by the same speaker:
\begin{align*}
\V{v}_{(c, q)} = [ \V{v}_c ; \V{v}_q ; \V{v}_c \circ \V{v}_q ; \phi(c, q) ]
\end{align*}
The antecedent score $f_a(c, q)$ is parameterized with $\V{W}_a$ and $\V{v}_a$ as follows:
\begin{align*}
f_a(c, q) = \V{v}_a \cdot \text{ReLU}(\V{W}_a \V{v}_{(c, q)})
\end{align*}

\paragraph{Pruning}
Holding the vector representation of every possible span in memory has a space complexity of $O(n^2 d)$ (where $n$ is the number of input tokens, and $d$ is the model's hidden dimension). This problem becomes even more acute when considering the space of span \textit{pairs} ($O(n^4 d)$).
Since this is not feasible, candidate mentions and antecedents are pruned through a variety of model-based and heuristic methods.

Specifically, mention spans are limited to a certain maximum length $\ell$.
The remaining mentions are then ranked according to their scores $f_m(\cdot)$, and only the top $\lambda n$ are retained, while avoiding overlapping spans.
Antecedents (span pairs) are further pruned using a lightweight antecedent scoring function (which is added to the overall antecedent score), retaining only a constant number of antecedent candidates $c$ for each target mention $q$.

\paragraph{Training}
For each remaining span $q$, the training objective optimizes the marginal log-likelihood of all of its unpruned gold antecedents $c$, as there may be multiple mentions referring to the same entity:
\begin{align*}
\log \sum_{c} P(a = c|q)
\end{align*}

\paragraph{Processing Long Documents}
Due to the c2f model's high memory consumption and the limited sequence length of most pretrained transformers, documents are often split into segments of a few hundred tokens each \cite{joshi-etal-2019-bert}.
Recent work on efficient transformers \cite{beltagy2020longformer} has been able to shift towards processing complete documents, albeit with a smaller model (base) and only one training example per batch.

\begin{figure*}[t]
\centering
\includegraphics[width=\linewidth]{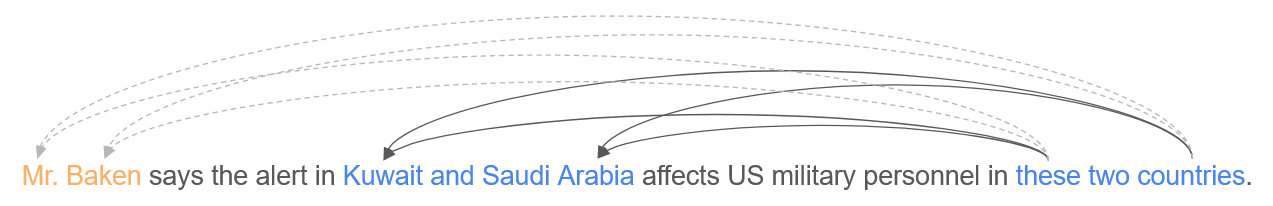}
\caption{The antecedent score $f_a(c,q)$ of a query mention $q=(q_s,q_e)$ and a candidate antecedent $c=(c_s,c_e)$ is defined via bilinear functions over the representations of their endpoints $c_s, c_e, q_s, q_e$. Solid lines reflect factors participating in positive examples (coreferring mentions), and dashed lines correspond to negative examples.}
\label{fig:illustration} 
\end{figure*}


\section{Model}\label{sec:model}

We present \textit{start-to-end} (s2e) coreference resolution, a simpler and more efficient model with respect to c2f (Section~\ref{sec:background}).
Our model utilizes the endpoints of a span (rather than all span tokens) to compute the mention and antecedent scores $f_m (\cdot)$ and $f_a (\cdot, \cdot)$ \textit{without} constructing span or span-pair representations; instead, we rely on a combination of lightweight bilinear functions between pairs of endpoint token representations.
Furthermore, our model does not use any handcrafted features, does not prune antecedents, and prunes mention candidates solely based on their mention score $f_m(q)$.

Our computation begins by extracting a \textit{start} and \textit{end} token representation from the contextualized representation $\V{x}$ of each token in the sequence:
\begin{align*}
\V{m}^{s} = \text{GeLU} (\V{W}_m^s \V{x}) \qquad
\V{m}^{e} = \text{GeLU} (\V{W}_m^e \V{x})
\end{align*}
We then compute each mention score as a biaffine product over the start and end tokens' representations, similar to \citet{dozat2016deep}:
\begin{align*}
f_m(q) = \V{v}_{s} \cdot \V{m}^{s}_{q_s} + \V{v}_{e} \cdot \V{m}^{e}_{q_e} + \V{m}^{s}_{q_s} \cdot \V{B}_m \cdot \V{m}^{e}_{q_e}
\end{align*}
The first two factors measure how likely the span's start/end token $q_s$/$q_e$ is a beginning/ending of an entity mention.
The third measures whether those tokens are the boundary points of the \textit{same} entity mention.
The vectors $\V{v}_{s}, \V{v}_{e}$ and the matrix $\V{B}_{m}$ are the trainable parameters of our mention scoring function $f_m$.
We efficiently compute mention scores for all possible spans while masking spans that exceed a certain length $\ell$.\footnote{While pruning by length is not necessary for efficiency, we found it to be a good inductive bias.}
We then retain only the top-scoring $\lambda n$ mention candidates to avoid $O(n^4)$ complexity when computing antecedents.

Similarly, we extract \textit{start} and \textit{end} token representations for the antecedent scoring function $f_a$:
\begin{align*}
\V{a}^{s} = \text{GeLU} (\mathbf{W}^s_a \V{x}) \qquad
\V{a}^{e} = \text{GeLU} (\mathbf{W}^e_a \V{x})
\end{align*}
Then, we sum over four bilinear functions:
\begin{align*}
f_a(c, q) &= \V{a}^{s}_{c_s} \cdot \V{B}_a^{ss} \cdot \V{a}^{s}_{q_s}
          + \V{a}^{s}_{c_s} \cdot \V{B}_a^{se} \cdot \V{a}^{e}_{q_e} \\
          &+ \V{a}^{e}_{c_e} \cdot \V{B}_a^{es} \cdot \V{a}^{s}_{q_s}
          + \V{a}^{e}_{c_e} \cdot \V{B}_a^{ee} \cdot \V{a}^{e}_{q_e}
\end{align*}
Each component measures the compatibility of the spans $c$ and $q$ by an interaction between different boundary tokens of each span.
The first component compares the \textit{start} representations of $c$ and $q$, while the fourth component compares the \textit{end} representations.
The second and third facilitate a cross-comparison of the \textit{start} token of span $c$ with the \textit{end} token of span $q$, and vice versa. Figure~\ref{fig:illustration} (bottom) illustrates these interactions.

This calculation is equivalent to computing a bilinear transformation between the concatenation of each span's boundary tokens' representations:
\begin{align*}
f_a(c, q) = [\V{a}^{s}_{c_s} ; \V{a}^{e}_{c_e}] \cdot \V{B}_a \cdot [\V{a}^{s}_{q_s} ; \V{a}^{e}_{q_e}]
\end{align*}
However, computing the factors \textit{directly} bypasses the need to create $n^2$ explicit span representations. Thus, we avoid a theoretical space complexity of $O(n^2 d)$, while keeping it equivalent to that of a transformer layer, namely $O(n^2 + nd)$.

\begin{table*}[t!]
\small
\centering
\begin{tabular}{@{}lcccccccccccccc@{}}
\toprule
\multicolumn{1}{c}{Model} & & \multicolumn{3}{c}{MUC} & & \multicolumn{3}{c}{B\textsuperscript{3}} & & \multicolumn{3}{c}{CEAF\textsubscript{$\phi_4$}}\\
 \cmidrule{1-1} \cmidrule{3-5} \cmidrule{7-9} \cmidrule{11-13}
   & & P & R & F1 & & P & R & F1 & & P & R & F1 & & Avg. F1 \\
\midrule
\midrule
c2f + SpanBERT-Large   & &    85.7 & \bf 85.3 &     85.5 & &     79.5 & \bf 78.7 & \bf 79.1 & & \bf 76.8 & 75.0 & 75.9 & & 80.2  \\
c2f + Longformer-Base  & &    85.0 &     85.0 &     85.0 & &     77.8 &     77.8 &     77.8 & & 75.6 & 74.2 & 74.9   & & 79.2  \\
c2f + Longformer-Large & &    86.0 &     83.2 &     84.6 & &     78.9 &     75.5 &     77.2 & & 76.7 & 68.7 & 72.5   & & 78.1 \\
\midrule
s2e + Longformer-Large & &\bf 86.5 & 85.1 & \bf 85.8   & & \bf 80.3 &    77.9 & \bf 79.1 & & \bf 76.8 & \bf 75.4 & \bf 76.1 & & \bf 80.3 \\ 
\bottomrule
\end{tabular}
\caption{Performance on the test set of the English OntoNotes 5.0 dataset. \textit{c2f} refers to the course-to-fine approach of \citet{lee-etal-2017-end, lee-etal-2018-higher}, as ported to pretrained transformers by \citet{joshi-etal-2019-bert}.}
\label{table:results}
\end{table*}

\begin{table}[t!]
\small
\centering
\begin{tabular}{@{}lcccc@{}}
\toprule
   & \textbf{Masc} & \textbf{Fem} & \textbf{Bias} & \textbf{Overall} \\
\midrule
\midrule
c2f + SpanBERT-Large   & 90.5 & \bf 86.3 & \bf 0.95 & \bf 88.4  \\
c2f + Longformer-Base  & 87.6 & 82.3 & 0.94 & 84.9  \\
c2f + Longformer-Large & 90.1 & 85.4 & \bf 0.95 & 87.8 \\
\midrule
s2e + Longformer-Large & \bf 90.6 & 85.8 & \bf 0.95 & 88.3 \\
\bottomrule
\end{tabular}
\caption{Performance on the test set of the GAP coreference dataset. The reported metrics are F1 scores.}
\label{table:gap}
\end{table}
\section{Experiments}
\label{section:experiments}


\paragraph{Dataset}
We train and evaluate on two datasets: the document-level English OntoNotes 5.0 dataset \cite{pradhan-etal-2012-conll}, and the GAP coreference dataset \cite{webster-etal-2018-mind}. 
The OntoNotes dataset contains speaker metadata, which the baselines use through a hand-crafted feature that indicates whether two spans were uttered by the same speaker.
Instead, we insert the speaker's name to the text every time the speaker changes, making the metadata available to any model.



\paragraph{Pretrained Model}
We use Longformer-Large \cite{beltagy2020longformer} as our underlying pretrained model, since it is able to process long documents without resorting to sliding windows or truncation.

\paragraph{Baseline}
We consider Joshi et al.'s \citeyearpar{joshi-etal-2019-bert} expansion to the c2f model as our baseline. 
Specifically, we use the implementation of \citet{xu-choi-2020-revealing} with minor adaptations for supporting Longformer.
We do not use higher-order inference, as \citet{xu-choi-2020-revealing} demonstrate that it does not result in significant improvements.
We train the baseline model over three pretrained models: Longformer-Base, Longformer-Large, and SpanBERT-Large \cite{beltagy2020longformer, joshi-etal-2020-spanbert}. 

\paragraph{Hyperparameters}
All models use the same hyperparameters as the baseline.
The only hyperparameters we change are the maximum sequence length and batch size, which we enlarge to fit as many tokens as possible into a 32GB GPU.\footnote{We made one exception, and tried to tune the Longformer-Large baseline's hyperparameters. Despite our efforts, it still performs worse than Longformer-Base.}
For our model, we use dynamic batching with 5,000 max tokens, which allows us to fit an average of 5-6 documents in every training batch.
The baseline, however, has a much higher memory footprint, and is barely able to fit a single example with Longformer-Base (max 4,096 tokens).
When combining the baseline with SpanBERT-Large or Longformer-Large, the baseline must resort to sliding windows to process the full document (512 and 2,048 tokens, respectively).


\paragraph{Performance}
Table \ref{table:results} and Table \ref{table:gap} show that, despite our model's simplicity, it performs as well as the best performing baseline. 
Our model with Longformer-Large achieves 80.3 F1 on OntoNotes, while the best performing baseline achieves 80.2 F1. 
When the baseline model is combined with either version of Longformer, it is not able to reach the same performance level as our model.
We see similar trends for GAP.
Our findings indicate that there is little to lose from simplifying the coreference resolution architecture, while there are potential gains to be had from optimizing with larger batches.


\paragraph{Efficiency}
We also compare our model's memory usage using the OntoNotes development set.
Table~\ref{tab:efficiency} shows that our implementation is at least three times more memory efficient than the baseline.
This improvement results from a combination of three factors: 
(1) the fact that our model is lighter on memory and does not need to construct span or span-pair representations, 
(2) our simplified framework, which does not use sliding windows, and 
(3) our implementation, which was written ``from scratch'', and might thus be more (or less) efficient than the original.

\begin{table}[t]
\small
\centering
\begin{tabular}{lr}
\toprule
Model & Memory (GB) \\
\midrule
c2f + SpanBERT-Large  & 16.2 \\
c2f + Longformer-Base  & 12.0 \\
c2f + Longformer-Large  & 15.7 \\
\midrule
s2e + Longformer-Large & ~~\textbf{4.3} \\
\bottomrule
\end{tabular}
\caption{Peak GPU memory usage during inference on OntoNotes, when processing one document at a time.}
\label{tab:efficiency}
\end{table}


\section{Related Work}

Recent work on memory-efficient coreference resolution sacrifices speed and parallelism for guarantees on memory consumption.
\citet{xia-etal-2020-incremental} and \citet{toshniwal-etal-2020-learning} present variants of the c2f model \cite{lee-etal-2017-end,lee-etal-2018-higher} that use an iterative process to maintain a fixed number of span representations at all times. Specifically, spans are processed sequentially, either joining existing clusters or forming new ones, and an eviction mechanism ensures the use of a constant number of clusters.
While these approach constrains the space complexity, their sequential nature slows down the computation, and slightly deteriorates the performance.
Our approach is able to alleviate the large memory footprint of c2f while maintaining fast parallel processing and high performance.

CorefQA \cite{wu-etal-2020-corefqa} propose an alternative solution by casting the task of coreference resolution as one of extractive question answering.
It first detects potential mentions, and then creates dedicated queries for each one, creating a pseudo-question-answering instance for each candidate mention.
This method significantly improves performance, but at the cost of processing hundreds of individual context-question-answer instances for a single document, substantially increasing execution time.
Our work provides a simple alternative, which can scale well in terms of both speed and memory.

\section{Conclusion}
We introduce a new model for coreference resolution, suggesting a lightweight alternative to the sophisticated model that has dominated the task over the past few years.
Our model is competitive with the baseline, while being simpler and more efficient.
This finding once again demonstrates the spectacular ability of deep pretrained transformers to model complex natural language phenomena.

\section*{Acknowledgements}

This research was funded by the Blavatnik Fund, the Alon Scholarship, Intel Corporation, and the Yandex Initiative in Machine Learning.

\bibliographystyle{acl_natbib}
\bibliography{anthology,acl2021}

\end{document}